\documentclass{article}

\usepackage{PRIMEarxiv}
\usepackage{microtype}
\usepackage{graphicx}
\usepackage{subfigure}
\usepackage{booktabs} 
\usepackage{xcolor}
\usepackage{physics}
\usepackage{amsfonts}
\usepackage{amssymb}
\usepackage{bbm}
\usepackage{cite}
\usepackage[ruled]{algorithm2e}
\usepackage{bm}
\usepackage{amsmath,amssymb,amsfonts}
\usepackage{algorithmic}

\usepackage[T1]{fontenc}

\usepackage[ruled]{algorithm2e}
\usepackage{bm}

\usepackage{longtable}
\usepackage{cite}
\usepackage{amsmath,amssymb,amsfonts}
\usepackage{algorithmic}
\usepackage{graphicx}
\usepackage{textcomp}
\usepackage{xcolor}
\usepackage{bm}
\def\BibTeX{{\rm B\kern-.05em{\sc i\kern-.025em b}\kern-.08em
    T\kern-.1667em\lower.7ex\hbox{E}\kern-.125emX}}

\usepackage{amsthm}
\usepackage{multirow}

\usepackage{caption}
\usepackage{url}            
\usepackage{booktabs}       
\usepackage{amsfonts}       
\usepackage{nicefrac}       
\usepackage{microtype}      
\usepackage{lipsum}
\usepackage{fancyhdr}       
\usepackage{graphicx}       
\graphicspath{{media/}}     

\usepackage{orcidlink}

\usepackage{hyperref}

\hypersetup{
  colorlinks=false,
  linkbordercolor=white,
 urlbordercolor=white,
pdfborder={0 0 0}
}

\title{LEAN-DMKDE: Quantum Latent Density Estimation for Anomaly Detection
\thanks{\textit{\underline{Citation}}: 
\textbf{Joseph et al., LEAN-DMKDE: Quantum Latent Density Estimation for Anomaly Detection.}} 
}

\author{
  Joseph A. Gallego-Mejia \orcidlink{0000-0001-8971-4998}, Oscar A. Bustos-Brinez \orcidlink{0000-0003-0704-9117} , Fabio A. Gonz\'{a}lez\orcidlink{0000-0001-9009-7288} \\
  MindLab \\
  Universidad Nacional de Colombia \\
  Bogot\'{a}, Colombia\\
  \texttt{\{jagallegom,oabustosb,fagonzalezo\}@unal.edu.co} 
}

\begin{document}
\maketitle

\begin{abstract}
This paper presents an anomaly detection model that combines the strong statistical foundation of density-estimation-based anomaly detection methods with the representation-learning ability of deep-learning models. The method combines an autoencoder, for learning a low-dimensional representation of the data, with a density-estimation model based on random Fourier features and density matrices in an end-to-end architecture that can be trained using gradient-based optimization techniques. The method predicts a degree of normality for new samples based on the estimated density. A systematic experimental evaluation was performed on different benchmark datasets. The experimental results show that the method performs on par with or outperforms other state-of-the-art methods.
\end{abstract}


\section{Introduction}

Anomaly detection is a critical task in several applications, such as fraud detection \cite{rizvi2020detection}, video surveillance \cite{Wang22video}, industrial defects \cite{denkena2020}, and medical image analysis \cite{medicalanomaly}, among others. In addition, it can be used as a preprocessing step in a machine learning system. In these scenarios, the idea is to detect whether a new sample is anomalous or not and leverage it to make actionable decisions \cite{Ruff2021}. Anomalies can change their name depending on the application domain, such as anomalies, outliers, novelties, exceptions, peculiarities, contaminants, among other terms. Anomaly detection refers to the task of finding patterns within the data that deviate from expected behavior \cite{aggarwal2016outlier}. 

Classical methods for dealing with anomaly detection fall into three main categories: classification-based such as one-class SVM \cite{scholkopf2001estimating}, distance-based such as local anomaly factor (LOF) \cite{breunig2000lof} and isolation forest \cite{liu2008isolation}, and statistically-based such as kernel density estimation \cite{nachman2020anomaly}. These methods have been used in several applications, however, they lack the representation learning ability of deep learning models that allow them to deal with complex, high-dimensional data such as images \cite{Liu2020}. Recent shallow methods, such as Stochastic Outlier Selection (SOS) \cite{Janssens2012} and COPOD \cite{Li2020copod}, are based on affinity and copulas, respectively. However, SOS constructs a matrix that is quadratic in terms of data points and COPOD does not directly compute joint distributions, which reduces its power. Recent deep learning methods, such as variational autoencoders (VAE) \cite{Pol19}, do not directly solve the anomaly detection problem, but solve an indirect task with the minimization of the reconstruction error. The VAE assumes that anomalous points have a higher value of the reconstruction error compared to normal points, however, the algorithm has been shown to have sufficient power to reconstruct even anomalous points. The deep support vector data description \cite{Ruff2018} surrounds the normal points in a hypersphere, but has an absence of degree of abnormality. Finally, the LAKE method \cite{Lv2020} is based on the union of two parts: variational autoencoder and kernel density estimation (KDE). One problem with this method is that it optimizes the variational autoencoder and feeds its results into the memory-based KDE algorithm. Each latent space in the training data set must be saved to feed the KDE, with its corresponding memory footprint. In this paper, we propose a new method, named LEAND, that uses the deep representation obtained by the autoencoders and the density captured by the density matrices, a formalism of quantum mechanics. The new method can be trained end-to-end, thus solving the shortcomings of LAKE.

The contributions of the present work are: 

\begin{itemize}
    \item A new algorithm to anomaly detection that can be trained end-to-end with current deep learning tools.
    \item A methodology for combining autoencoders with kernel methods to produce normality likelihood estimation.
    \item A systematic comparison on multiple datasets of the novel method against state-of-the-art anomaly detection methods.
\end{itemize}

\textbf{Reproducibility}: all the code is available at the Github repo http:// and a release in Zenodo. 

\section{Background and Related Work}

Anomaly detection is a well studied topic in machine learning. The idea is to detect unexpected behavior which deffer from the normal, expected behavior. Given $r$ as the proportion of anomalous data points, a proportion greater than 10\% is typically solved using a classification based approach such as One Class Support Vector Machines \cite{scholkopf2001estimating}. Nonetheless, in general, anomalous data points are scarce and datasets are unbalanced, with a proportion of anomalous data points commonly less than 2\%. The two main assumptions here are that the anomalous points are scarce and the anomalous points deviates greatly from the normal points.      

In recent years, classical models have been progressively replaced by deep learning algorithms, showing high performance in anomaly detection tasks \cite{Zhang2021}. Deep algorithms can model the features of complex data and perform feature engineering, especially if a large amount of data is available to process \cite{Rippel2021}. For this, they are commonly used in scenarios involving very large datasets with high-dimensional instances, such as images and video.

\subsection{Baseline Anomaly Detection Methods}

\subsubsection*{Classical Methods}
\begin{itemize}
    \item Isolation Forest \cite{liu2008isolation}: this method profiles the anomalous data points from the beginning, assuming that some anomalous points can be found in the training data set. It divides the space into lines orthogonal to the origin. The points that require less tree breaking are considered anomalous. 
    \item Minimum Covariance Determinant  \cite{rousseeuw1999fast}: this algorithm embeds the data points into a hyperellipsoid using robust estimators of the mean and covariance of the points, and then uses the Mahalanobi distance of each point to score its anomaly. 
    \item One Class Support Vector Machines \cite{scholkopf2001estimating}: the method generates a maximal margin from the origin such that all the training data points (normal points) lie within the origin and the margin. The method suppose that anomalous points are far away from normal points, i.e, they resides in the subspace generated by the maximization of the hyperplane that does not contain the origin. A Gaussian Kernel was used for all the experiments given their good properties. 
    \item Local Outlier Factor \cite{breunig2000lof}: the algorithm computes the $k$-nearest neighbors of each point. It then computes the local reach-ability density using the distance between the points and the $k$-nearest neighbor. Next, the algorithm compares the local reach-ability density of each point and compute the LOF. When the LOF values is equal or less than 1, the data point is considered an inlier, otherwise an outlier. 
    \item K-nearest neighbors (KNN) \cite{Ramaswamy2000}: this method calculates the $k$-nearest neighbor distances of each point and computes a function over this distance. The three main functions to compute the distances are: the largest distance, the mean distance and the median distance. According to the distance computed function, the scientist selects an anomaly threshold given the percentage of anomalous data points.
\end{itemize}

\subsubsection*{Recent Shallow Methods}
\begin{itemize}
    \item SOS \cite{Janssens2012}: the algorithm is inspired by T-SNE. It computes a dissimilarity matrix which can be the Euclidean distance. The dissimilarity matrix is then used to compute an affinity matrix using a perplexity parameter. With stochastic selection of sub-graphs, a probability metric of being an outlier point is calculated using the link probabilities between points in the sub-graph. Given the link probability, a threshold value is selected to classify outlier points from normals. 
    \item COPOD \cite{Li2020copod}: this algorithm uses the copula, which is a function that allows the joint multivariate distribution function to be expressed in its marginal distribution functions. The copula is calculated and then a skewness correction is calculated to decide whether to use the left tail or the right tail of the copula. Finally, a threshold value is used to determine whether a point is anomalous or normal. 
    \item LODA \cite{Pevny2016}: it is an ensemble algorithm based on histograms applied to each dimension. Each histogram is projected using $\bm{w}$ random parameters. The joint probability is computed under the strong assumption of independence between the projected vectors. The logarithm function is applied to avoid the curse of dimensionality.
\end{itemize}

\subsubsection*{Deep Learning Methods}
\begin{itemize}
    \item VAE-Bayes \cite{an2015variational}: this model is based on Bayesian statistics. It uses an autoencoder with stochastic variational inference to reconstruct the original sample point. 
    A high reconstruction error for a given sample is considered an anomalous data point.
    \item DeepSVDD \cite{Ruff2018}: this method is based on Support Vector Data Description (SVDD), which surrounds normal points on a hypersphere in a reproducing Hilbert space. The anomalous points are assumed to be outside the hypersphere. The parameter $v$ controls the proportion of anomalous points found by the algorithm. 
    \item LAKE \cite{Lv2020}: it is based on the union of two anomaly detection methods. First, a variational autoencoder (VAE) is used as a dimensionality reduction algorithm and as a measure of reconstruction error. Second, a kernel density estimation (KDE) is attached to the last hidden layer of the variational autoncoder and concatenate with the reconstruction error in terms of mean squared error and cosine similarity. The KDE use the reconstruction error and the dimensionality reduction to construct an estimate of the density of the given point. 
\end{itemize}

\section{Quantum Latent Density Estimation for Anomaly Detections (LEAND)} \label{sec:leand}

\begin{figure*}[t]
\begin{centering}
\includegraphics[scale=0.6]{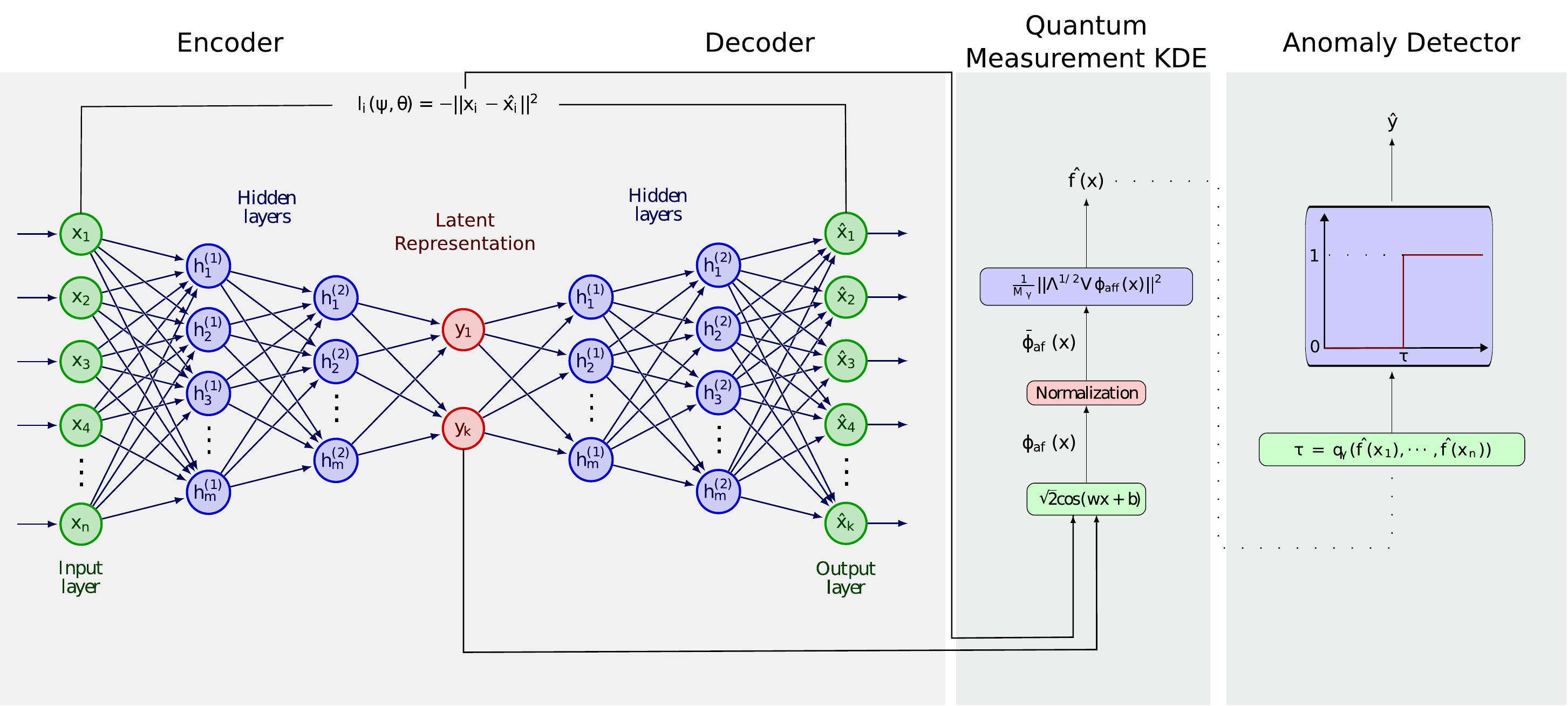}
\par\end{centering}
\caption{Quantum Anomaly Detection through Density Matrices, Adaptive Fourier Features and Autoencoders (LEAND) method. Step 1, autoencoder representation learning. Step 2, training of the model using both the reconstruction error and the maximum likelihood estimation of the density matrix $\rho$. Step 3, estimation of the density of a new sample using the learn density matrix $\rho=V^T\Lambda V$. Step 4, anomaly detector using the proportion of the anomalies.   \label{fig:model}}
\end{figure*}

The proposed method Quantum Latent Density Estimation for Anomaly Detection (LEAND) is shown in Figure \ref{fig:model}. This model is composed of an autoencoder, an adaptive Fourier feature layer, a quantum measurement layer and finally a threshold anomaly detection layer. The autoencoder is responsible for performing a dimensionality reduction and calculating the reconstruction error. The adaptive Fourier feature layer maps the autoencoder latent space to an Hilbert space whose dot product approximates the Gaussian kernel. The quantum measurement layer produces density estimates of the given data points. The last step of the algorithm is an anomaly detector that uses the density estimate and classifies each sample as anomalous or normal.

\subsection{Autoencoder}

The first step of the algorithm is an autoencoder. Given an input space $\mathbb{R}^d$, and a latent space $\mathcal{R}^p$, where typically $p \ll d$, a point in the input space is sent through an $\psi$-encoder and a $\theta$-decoder step, where $\psi: \mathcal{R}^d \rightarrow \mathcal{R}^p$ and $\theta: \mathcal{R}^p \rightarrow \mathcal{R}^d$,  such that:

$$\bm{z}_i = \psi(\bm{x}_i, \bm{w}_\psi)$$
$$\hat{\bm{x}_i} = \theta(\bm{z}_i, \bm{w}_\theta)$$

where $\bm{w}_\psi$ and $\bm{w}_\theta$ are the network parameters of the encoder and the decoder respectively, and $\hat{x}$ is the reconstruction of the original data. The reconstruction error of the autoencoder is defined as the Euclidean distance between the original data point and its reconstruction:

\begin{align} \label{eq:encoder_loss}
\begin{aligned}
l_i(\psi, \theta) = -||\bm{x}_i - \hat{\bm{x}_i}||^2
\end{aligned}
\end{align}

To capture further information computed by the autoencoder, the Euclidean distance error ($l_{\text{Euc\_dist},i}$) were combined with the cosine similarity ($l_{\text{cos\_sim},i}$), defined as the cosine of the angle between $\bm{x}_i$ and $\hat{\bm{x}_i}$, into one vector with the latent space ($\bm{z}_i$) as:

$$\text{rec\_measure}_i = [l_{\text{Euc\_dist},i}, l_{\text{cos\_sim},i}]$$ 
and the latent space output of the autoencoder to the next layer is $\bm{o}_i = [\bm{z}_i, \text{rec\_measure}_i]$.

\subsection{Quantum Measurement Kernel Density Estimation}

Density estimation is one of the most studied topics in statistics. Given a random variable $X$ and its associated measurable space $\mathcal{A}$, from which we capture some events, observations or samples $x_1, \cdots, x_n$, the probability density function $f$ in a measurable set $A \in \mathcal{A}$ is defined as a function that satisfies the following property: $$Pr[X\in A]  = \int_A f_X(\bm{x}) d\bm{x}$$

In general, the underlying density function of an arbitrary system is unknown. The most popular method for density estimation is kernel density estimation (KDE), also known as Parzen window. KDE has two principal drawbacks: its memory-based approach, i.e., it stores each data point $x_1, \cdots, x_n$ in the training phase and uses them to produce inferences in the testing phase, and it cannot be combined with deep learning architectures. Given a set of training points $\{\bm{x}_i\}_{i=1,\cdots,N}$, and a point $\bm{x}$ whose density is to be calculated, the Kernel Density Estimation calculates it as: 

\begin{equation}
    \hat{f}(\bm{x})= \frac{1}{N M_\gamma} \sum_{i=1}^N k_\gamma({\bm{x}_i},\bm{x})
    \label{eq:kernel_density_estimation}
\end{equation}
where $\gamma$ is the bandwidth parameter, $M\gamma$ is a normalization parameter, and $k_\gamma$ is a kernel with specific properties \cite{girolami03}.   

In \cite{gonzalez2021learning, josephQuantumNeural}, the authors showed that using a derivation of the kernel density estimation method using the Gaussian kernel, we can derive a density estimation method that captures quantum information as follows: 

\begin{align} \label{eq:kde_density_matrix}
\begin{aligned}
\hat{f}(\bm{x}) & = \frac{1}{M_\gamma} {\phi}_\text{aff}(\bm{x})^T \rho \: {\phi}_\text{aff}(\bm{x})
\end{aligned}
\end{align}
affwhere $\rho$ is a density matrix defined as $$ \rho = \frac{1}{N}\sum_{i=1}^N \phi_\text{aff}({\bm{x}_i})\phi^T_\text{aff}({\bm{x}_i})$$ and $M_\gamma$ is a normalization constant. Note that $\rho$ is a $D$-dimensional square matrix. The density matrix captures the classical and quantum probabilities of a quantum system, and it can be viewed as an aggregation of each data point in the feature space. The $\rho$ matrix is required to be a Hermitian matrix and its trace must be equal to one, $tr(\rho)=1$.   

$\hat{\phi}_\text{aff}$ indicates the adaptive Fourier features, which are explained right next. Random Fourier features were first proposed by \cite{rahimi2007rff}. Its main idea is to build an explicit approximate mapping of the reproducing Hilbert space induced by a given kernel. In particular, they showed that if a Gaussian kernel is used, we can approximate $$k(\bm{x}, \bm{y}) \simeq \mathbb{E}_{\bm{w}}(\langle \hat{\phi}_{\text{aff}, \bm{w}}(\bm{x}), \hat{\phi}_{\text{aff}, \bm{w}}(\bm{y})\rangle )$$
where $\hat{\phi}_{\text{aff}, \bm{w}} = \sqrt{2} \cos (\bm{w}^T\bm{x} + b)$,  $\bm{w} \sim N(\bm{0},\mathbb{I}^d)$ and $\bm{b} \sim \text{Uniform}(0, 2\pi)$. This expectation approximation is possible due to Bochner's theorem. The adaptive Fourier features are a neural network-based fine-tuning of the random Fourier features. The explanation of this network and its usage is presented in the Training Strategy subsection. \ref{subsect:training_strategy}

The dimensions of the $\rho$ matrix can be problematic, given its size $D^2$, but this can be alleviated by using the matrix eigen-decomposition:

\begin{equation}
    \rho \approx V^T \Lambda V
\end{equation}
where $V \in \mathcal{R}^{r \times D}$ is a matrix whose rows contain the $r$ vectors with the largest eigenvalues, and $\Lambda \in \mathcal{R}^{r \times r}$ is a diagonal matrix whose values correspond to the $r$ largest eigenvalues of $\rho$. This matrix eigen-decomposition modify the Equation \ref{eq:kde_density_matrix} to:

\begin{align} \label{eq:AFDM_estimation}
\begin{aligned}
\hat{f}_\gamma(\bm{x})& \simeq \frac{1}{M_\gamma}|| \Lambda^{1/2}V{\phi}_\text{aff}(\bm{x})||^2
\end{aligned}
\end{align}

Equation \ref{eq:AFDM_estimation} defines the algorithm to perform new density estimates based on density matrices. The parameters $\Lambda$ and $V$ are obtained by gradient descent optimization. Note that the adaptive Fourier features are trained independently using only the autoencoder and adaptive Fourier feature layer. The output of the adaptive Fourier feature layer is the input of the density matrices using the Equation \ref{eq:AFDM_estimation}. The log likelihood optimization process is as follows:

$$V^*, \Lambda^* = \arg \min_{V, \Lambda} \alpha \sum_{i=1}^N \log \hat{f}(\bm{x}_i)$$

 where $\hat{f}(\bm{x}_i) =\frac{1}{M_\gamma}|| \Lambda^{1/2}V\phi_\text{aff}(\bm{x})||^2$  and $\phi_{\text{aff}}(\bm{x}_i)= \sqrt{2}cos(\bm{w}^T\bm{x}_i + b)$. The algorithm is trained minimizing both reconstruction and estimation losses using gradient descent.

\subsection{Anomaly Detection Step}

The last step of the anomaly detection algorithm is to detect whether a given data point is a normal or anomalous point. The algorithm uses the proportion of anomalies ($\gamma$) within the data set to obtain a threshold value $\tau$ as:

$$\tau = q_{\gamma} (\hat{f}(\bm{x}_1), \cdots, \hat{f}(\bm{x}_n))$$
where $q$ is the percentile function. In the detection phase, the algorithm uses the $\tau$ value to classify anomalous points as:  
$$\hat{y}(x_{i}) = \left\{\begin{matrix}
 \text{ `normal'} & \textit{if } \hat{f}(x_{i}) \geq \tau\\ 
 \text{ `anomaly'} & \textit{otherwise}
\end{matrix}\right.$$

\subsection{Training Strategy \label{subsect:training_strategy}}

The random Fourier feature approximation can be further tuned using gradient descent, in a process called ``Adaptive Fourier features". This algorithm was first proposed in \cite{josephQuantumNeural}. It selects random pairs of data points $(\bm{x}_i, \bm{x}_j)$ and applies gradient descent to find:

$$\bm{w}^*, b^* = \arg \min_{\bm{w},b} \frac{1}{m}\sum_{\bm{x}_i,\bm{x}_j\in s} (k_\gamma(\bm{x}_i,\bm{x}_j) - \hat{k}_{\bm{w},b}(\bm{x}_i,\bm{x}_j))^2$$
where $k_\gamma(\bm{x}_i,\bm{x}_j)$ is the Gaussian kernel with a $\gamma$-bandwidth parameter, and $\hat{k}_{\bm{w},b}(\bm{x}_i,\bm{x}_j))^2$ is the Gaussian kernel approximation. As an initialization to the Gaussian kernel approximation, we can sample $\bm{w}$ and $b$ using $N(\bm{0},\mathbb{I}^d)$, and $\text{Uniform}(0, 2\pi)$, respectively, the same as the random Fourier features. 

It should be noted that each $\bm{x}_i$ is a vector with $m$ dimensions. For the anomaly detection algorithm, $m$ is the latent space of the autoencoder defined above. Therefore, we propose an intermediate step to approximate the kernel in the latent space. First, the autoencoder is trained without any additional layer. Second, we use forward propagation for several $\bm{x}_i$ data points to reduce the original space to the encoded space. Finally, the adaptive Fourier feature algorithm is optimized as proposed above.

The training process is shown in the Algorithm \ref{alg:algorithm}. The hyperparameter $\alpha$ controls the trade-off between the minimization of the reconstruction error and the maximization of the log-likelihood. Using forward-propagation, each $\bm{x}_i$ is send through the encoder $\psi(\bm{x}_i, \bm{w}_\psi)$ and the decoder $ \theta(\bm{z}_i, \bm{w}_\theta)$. Then, reconstruction errors are computed and combined with the encoded $\bm{z}_i$ as $\bm{o}_i$. Using the parameters of the adaptive Fourier features, the vector $\bm{o}_i$ is mapped to a Hilbert space in which dot product approximates the Gaussian kernel. With the mapping $\phi_{\text{aff}, \bm{w}_{\text{aff}}}$, we compute the likelihood of each sample using the density matrix decomposition obtaining $\hat{f}(\bm{x}_i)$. A minimization optimization is performed using backpropagation and stochastic gradient descent. Finally, $\tau$ is computed as the percentile associated to a given $\gamma$ proportion of anomalies.

\begin{algorithm}[tb]
\caption{LEAND training process}
\label{alg:algorithm}
\textbf{Input}: Training dataset $D=\{\bm{x}_i\}_{i=1,\cdots,N}$ \\
\textbf{Parameters}: $\alpha$: trade-off between reconstruction error and log-likelihood,\\
$\bm{w}_{\text{rff}},b$: adaptive Fourier features parameters\\
$\gamma$: proportion of anomalies\\
\textbf{Output}: $\bm{w}_\psi$: encoder parameters , \\
$\bm{w}_\theta$: decoder parameters, \\
$V^*, \Lambda^*:$ quantum measurement KDE parameters \\
\begin{algorithmic}[1] 
\FOR{$\bm{x}_i \in D \textbf{ until convergence}$}
\STATE $\bm{z}_i = \psi(\bm{x}_i, \bm{w}_\psi)$
\STATE $\hat{\bm{x}_i} = \theta(\bm{z}_i, \bm{w}_\theta)$
\ENDFOR 
\STATE $\bm{w}_{\text{AFF}}^*, b_{\text{AFF}}^* = \arg \min_{\bm{w},b} \frac{1}{m}\sum_{\bm{x}_i,\bm{x}_j\in D}( k_{\gamma}(\psi_{\bm{w}_\psi}(\bm{x}_i),\psi_{\bm{w}_\psi}(\bm{x}_j)-\hat{k}_{\bm{w},b}(\psi_{\bm{w}_\psi}(\bm{x}_i),\psi_{\bm{w}_\psi}(\bm{x}_j)))^2$
\FOR{$\bm{x}_i \in D \textbf{ until convergence}$}
\STATE $\bm{z}_i = \psi(\bm{x}_i, \bm{w}_\psi)$
\STATE $\hat{\bm{x}_i} = \theta(\bm{z}_i, \bm{w}_\theta)$
\STATE $l_{\text{Euc\_dist},i}=\text{Euclidean\_distance}(\bm{x}_i,\bm{\hat{x}}_i)$
\STATE $l_{\text{cos\_sim},i}=\text{cosine\_similarity}(\bm{x}_i,\bm{\hat{x}}_i)$
\STATE $\bm{o}_i = [\bm{z}_i,l_{\text{Euc\_dist},i}, l_{\text{cos\_sim},i}]$
\STATE  $\hat{f}(\bm{o}_i) =\frac{1}{M_\gamma}|| \Lambda^{1/2}V\phi_{\text{aff},\bm{w}}(\bm{\bm{o}_i})||^2$  
\STATE $\mathbf{L}_{\bm{w}_\psi,\bm{w}_\theta,V^*, \Lambda^*} = ||\bm{x}_i - \hat{\bm{x}_i}||^2 - \alpha \sum_{i=1}^N \log \hat{f}(\bm{o}_i)$
\STATE $\bm{w}_\psi,\bm{w}_\theta,V^*, \Lambda^* \leftarrow \text{update parameters minimizing } \mathbf{L}$
\ENDFOR 
\STATE $\tau = q_{\gamma} (\hat{f}(\bm{x}_1), \cdots, \hat{f}(\bm{x}_n))$

\STATE \textbf{return} $\bm{w}_\psi,\bm{w}_\theta,V^*, \Lambda^*, \tau$
\end{algorithmic}
\end{algorithm}

\section{Experimental Evaluation}

\subsection{Experimental Setup}

The experimental framework used in this paper intends to compare LEAND with all the baseline algorithms listed in previous sections. To run One Class SVM, Minimum Covariance Determinant, Local Outlier Factor and Isolation Forest, the implementation used is the one provided by Scikit-Learn Python library. KNN, SOS, COPOD, LODA, VAE and DeepSVDD were run using the implementation provided by PyOD Python library \cite{Zhao2019}. LAKE algorithm was implemented based on the Github repository of its authors, although we had to correct the way the test dataset were split to include both normal and anomalous samples in it.

In order to handle the inherent randomness found in some of the algorithms, it was decided to fix in advance (into a single, invariant number) all the random seeds that could affect the different stages of each algorithm (particularly dataset splitting and initialization steps). All experiments were carried out on a machine with a 2.1GHz Intel Xeon 64-Core processor with 128GB RAM and two RTX A5000 graphic processing units, that run Ubuntu 20.04.2 operating system.

\subsubsection{Datasets}

To evaluate the performance of LEAND for anomaly detection tasks, twenty public datasets were selected. These datasets came from two main sources: the Github repository associated with LAKE\footnote{https://github.com/1246170471/LAKE}, and the ODDS virtual library of Stony Brook University\footnote{http://odds.cs.stonybrook.edu/about-odds/}. The main characteristics of the selected datasets, including their sizes, dimensions and outlier rates can be seen in Table \ref{dataset_features}. These datasets were chosen because of the wide variety of features they represent, with which the performance of the algorithms can be tested in a multiple scenarios, its extensive use in anomaly detection literature, and because of their accessibility, since the files associated with each dataset can be easily accessed in their respective sources. 

\begin{table}[t]
\centering
\small
\begin{tabular}[c]{l|c|c|c}
\hline
Dataset & Instances & Dimensions & Outlier Rate \\ 
\hline

Arrhythmia & 452 & 274 & 0,146\\
Cardio & 2060 & 22 & 0,2\\
Glass & 214 & 9 & 0,042 \\
KDDCUP & 5000 & 118 & 0,1934 \\
Lympho & 148 & 18 & 0,04 \\
Ionosphere & 351 & 33 & 0,359\\
Letter & 1600 & 32 & 0,0625 \\
MNIST & 7603 & 100 & 0,092 \\
Musk & 3062 & 166 & 0,0317  \\
OptDigits & 5216 & 64 & 0,0288 \\
PenDigits & 6870 & 16 & 0,0227\\
Pima & 768 & 8 & 0,349 \\
Satellite & 6435 & 36 & 0,3164 \\
SatImage & 5803 & 36 & 0,0122 \\
Shuttle & 5000 & 9 & 0,0715 \\
SpamBase & 3485 & 58 & 0,2  \\
Thyroid & 3772 & 36 & 0,0247  \\
Vertebral & 240 & 6 & 0,125 \\
Vowels & 1456 & 12 & 0,03434  \\
WBC & 378 & 30 & 0,0556  \\  
\hline
\end{tabular}

\caption{Main features of the datasets.}
\label{dataset_features}
\end{table}

\subsubsection{Metrics}

To compare the performance of the proposed algorithm against the baseline anomaly detection methods, F1-Score (with weighted average) was selected as the main metric. This is a widely used metric in machine learning frameworks. However, the calculation of other metrics, such as accuracy, F1-Score exclusive for anomaly class, the area under the Recall-Precision curve (AUC-PR) and the area under the ROC curve (AUC-ROC), were also computed but not reported. The specific calculations of F1-Score and the other additional metrics follow the implementation provided in Scikit-Learn Python library \cite{scikit-learn}.

For each algorithm, a set of parameters of interest was selected in order to perform a series of searches for the combinations of parameters that gave the best results for each dataset under study. The list of algorithms and their parameters under study can be found in Table \ref{parameters}.

\begin{table}[ht!]
\small
\centering
\begin{tabular}[c]{p{0.12\textwidth}|p{0.31\textwidth}}
\hline
Algorithm & Parameters of Interest \\
\hline

One Class SVM & gamma (Gaussian kernel), outlier percentage \\
Isolation Forest & number of estimators, samples per estimator, outlier percentage \\
Covariance Estimator & outlier percentage \\
LOF (Local Outlier Factor) & number of neighbors, outlier percentage \\
K-nearest Neighbors & number of neighbors, outlier percentage \\
SOS & perplexity (matrix parameter), outlier percentage \\
COPOD & outlier percentage \\
LODA & outlier percentage \\
VAE-Bayes & outlier percentage \\
DeepSVDD & outlier percentage \\
LAKE & normality ratio (related to outlier percentage) \\
LEAND & sigma (Gaussian kernel), architecture of the autoencoder, size of Fourier features mapping, size of density matrix eigen-decomposition, alpha (trade-off parameter)  \\

\hline
\end{tabular}

\caption{Parameters of all the algorithms selected for grid search.}
\label{parameters}
\end{table}

The selection of the best parameters was made by using a grid search strategy, ranking all the possible combinations of parameters (up to a limit of 100 experiments) in terms of the weighted F1-Score, and choosing the combination that showed the highest value for the metric. This selection of parameters was performed for each algorithm and each dataset.

\subsection{Results and Discussions}

\begin{table*}[ht!]
\small
\begin{tabular}{l||c|c|c|c|c|c|c|c|c|c|c||c}
\hline
Dataset    & OCSVM & iForest & \tiny{Covariance} & LOF & KNN & SOS & COPOD & LODA   & \tiny{VAE-Bayes} & \tiny{DeepSVDD} & LAKE   & LEAND \\
\hline
Arrhythmia & 0,813       & 0,821   & 0,818      & 0,804 & 0,861 & 0,773 & 0,844 & 0,798 & 0,856     & 0,864    & 0,849 & \bf{0,902}      \\
Cardio     & 0,804       & 0,752   & 0,756      & 0,702 & 0,753 & 0,739 & 0,750 & 0,717 & 0,783     & 0,735    & 0,835 & \bf{0,856}      \\
Glass      & 0,916       & 0,931   & 0,931      & 0,925 & 0,900 & 0,908 & 0,916 & 0,848 & 0,900     & 0,908    & \bf{1,000} & \bf{1,000}      \\
Ionosphere & 0,765       & 0,710   & 0,876      & 0,830 & 0,817 & 0,784 & 0,736 & 0,510 & 0,714     & 0,674    & 0,934 & \bf{0,973}      \\
KDDCUP     & 0,744       & 0,838   & 0,975      & 0,789 & 0,932 & 0,716 & 0,785 & 0,770 & 0,803     & 0,755    & 0,982 & \bf{0,991}      \\
Letter     & 0,893       & 0,897   & 0,909      & 0,930 & 0,910 & 0,911 & 0,895 & 0,899 & 0,897     & 0,897    & \bf{0,962} & 0,919      \\
Lympho     & 0,934       & \bf{1,000}   & 0,934      & \bf{1,000} & \bf{1,000} & 0,934 & 0,962 & 0,923 & \bf{1,000}    & \bf{1,000}   & \bf{1,000} & \bf{1,000}      \\
MNIST      & 0,881       & 0,881   & 0,841      & 0,886 & 0,909 & 0,864 & 0,868 & 0,866 & 0,895     & 0,880    & \bf{0,970} & 0,930      \\
Musk       & 0,958       & 0,931   & 0,997     & 0,958 & 0,991 & 0,951 & 0,964 & 0,954 & 0,984     & 0,992    & 0,991 & \bf{1,000}     \\
OptDigits  & 0,952       & 0,952   & 0,952      & 0,949 & 0,953 & 0,953 & 0,949 & 0,955 & 0,951     & 0,952    & \bf{0,977} & 0,958     \\
PenDigits  & 0,963       & 0,967   & 0,966      & 0,961 & 0,965 & 0,963 & 0,966 & 0,966 & 0,966     & 0,963    & \bf{0,994} & 0,989      \\
Pima       & 0,592       & 0,624   & 0,558      & 0,615 & 0,644 & 0,636 & 0,615 & 0,597 & 0,632     & 0,679    & \bf{0,874} & 0,742      \\
Satellite  & 0,681       & 0,757   & 0,813      & 0,634 & 0,716 & 0,597 & 0,732 & 0,709 & 0,761     & 0,761    & \bf{0,921} & 0,852     \\
SatImage   & 0,984       & 0,999   & 0,991      & 0,979 & 0,998 & 0,980 & 0,994 & 0,997 & 0,996     & 0,996    & \bf{1,000} & \bf{1,000}      \\
Shuttle    & 0,933       & 0,992   & 0,968      & 0,900 & 0,973 & 0,891 & 0,990 & 0,973 & 0,981     & 0,978    & 0,985 & \bf{0,996}      \\
SpamBase   & 0,702       & 0,794   & 0,714      & 0,702 & 0,719 & 0,719 & 0,799 & 0,699 & 0,741     & 0,738    & 0,850 & \bf{0,947}      \\
Thyroid    & 0,953       & 0,958   & \bf{0,986}      & 0,949 & 0,953 & 0,949 & 0,953 & 0,958 & 0,958     & 0,960    & 0,874 & 0,972     \\
Vertebral  & 0,750       & 0,778   & 0,817      & 0,796 & 0,810 & 0,817 & 0,817 & 0,817 & 0,817     & 0,817    & 0,807 & \bf{0,870}      \\
Vowels     & 0,950       & 0,942   & 0,941      & 0,951 & 0,969 & 0,954 & 0,943 & 0,929 & 0,952     & 0,943    & 0,949 & \bf{0,986}      \\
WBC        & 0,942       & 0,941   & 0,949      & 0,957 & 0,942 & 0,913 & 0,970 & 0,947 & 0,957     & 0,957    & \bf{1,000} & 0,972     
\end{tabular}
\caption{F1-Score for all classifiers over all datasets.}
\label{f1_score}
\end{table*}

The F1-Scores obtained when using the winning combinations for all experiments (algorithm and dataset pairs) are reported in Table \ref{f1_score}. At a first glance, there is a noticeable difference between the performance of most of the baseline methods and LEAND, with LAKE being a notable exception; other methods that show better-than-average results include Covariance and KNN, that had a slightly better performance than deep learning alternatives like VAE-Bayes or DeepSVDD.

When considering the influence of the dataset features over the performance, some patterns emerge. With respect to the outlier rate of the datasets, LEAND performs well for datasets whose outlier rate lie below 10\%, but stands the most in datasets between 10\% and 30\%. There is also a less notable relationship with the size of the dataset (LEAND performs slightly better for datasets of sizes between 1000 and 5000 samples) and with the dimension of data points, where LEAND stands more in datasets with big dimensionality (more than 50 dimensions).

\begin{figure}
    \centering
    \includegraphics[width=0.42\textwidth]{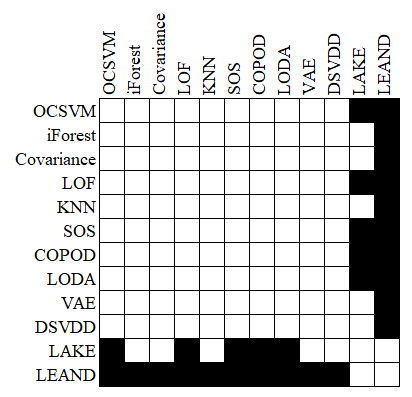}
    \caption{Results for Friedman-Nemenyi test over the metrics. Our method (LEAND) is the most different to others, followed by LAKE.}
    \label{fig:chess_nemenyi}
\end{figure}

Over these results, a statistical analysis was performed. In particular, the Friedman test was applied over F1-Scores to conclude that there is a statistically significant difference between the methods; then, a Friedman-Nemenyi test was also applied. The results of this latter test can be seen in Figure \ref{fig:chess_nemenyi}, where black squares correspond to pairs of datasets that differ significantly, and white squares correspond to pairs that do not. LEAND stands out as the most different with respect to all other methods, where LAKE is the only other method that differs of some other methods.

\subsection{Ablation Study}

To establish that the LEAND architecture can achieve a better performance than each of its separate components, the following experiments were performed: 

\begin{itemize}
    \item KDE only: density estimation is at the core of the proposed method, so a classic approach to KDE (specifically, the one from Scikit-Learn library) was also applied over the datasets. It was allowed to operate with different kernel functions and bandwidths.
    \item AutoEncoder only: since the proposed method uses an autoencoder as a mechanism to create a latent representation of the data, the neural network alone was applied over the datasets, in such a way that the reconstruction error was used as a measure of anomaly.
    \item Removing reconstruction measure: LEAND uses both reconstruction measure (from the autoencoder) and estimation error (from the quantum estimation stage) as the loss function that it tries to minimize. A LEAND version (called LEAND NoRecon) where only the estimation measure is considered as the loss function was also applied on the datasets.
\end{itemize}

The performance of these different experiments over all datasets are compared with LEAND using F1-Score as before. A parameter grid search was also performed over the parameters that applied in each case and with the same value ranges used in LEAND. Results can be seen in Table \ref{ablation}. 

From these results, it is posible to see that LEAND performs effectively better than isolating KDE or the Autoencoder stages, mostly with the latter. KDE show good performances in some datasets, possibly due to the fact that it was allowed to work with different kernel shapes, some of which can fit better to certain datasets than the Gaussian kernel that LEAND utilizes. Autoencoder alone and LEAND NoRecon do not seem to perform as well as LEAND (except in a few cases), showing that relying only on reconstruction error is a notable shortcoming when compared to combining it with density estimation.

\begin{table}[ht!]
\centering
\small
\begin{tabular}[c]{l||c|c|c||c}
\hline
Dataset     & KDE   & AE & \tiny{LEAND NoRecon} & LEAND \\
\hline
Arrhythmia  & 0,895 & 0,865 & 0,887 & \bf{0,902} \\
Cardio      & 0,847 & 0,809 & 0,847 & \bf{0,856} \\
Glass       & 0,934 & 0,939 & 0,974 & \bf{1,000} \\
Ionosphere  & 0,946 & 0,814 & 0,959 & \bf{0,973} \\
KDDCUP      & 0,989 & 0,857 & 0,988 & \bf{0,991} \\
Letter      & \bf{0,929} & 0,907 & 0,914 & 0,919 \\
Lympho      & \bf{1,000} & \bf{1,000} & \bf{1,000} & \bf{1,000} \\
MNIST       & 0,925 & 0,883 & 0,918 & \bf{0,930} \\
Musk        & \bf{1,000} & 0,981 & \bf{1,000} & \bf{1,000} \\
OptDigits   & \bf{0,986} & 0,959 & 0,960 & 0,958 \\
PenDigits   & \bf{0,996} & 0,968 & 0,989 & 0,989 \\
Pima        & 0,734 & 0,708 & 0,733 & \bf{0,742} \\
Satellite   & 0,821 & 0,689 & 0,848 & \bf{0,852} \\
SatImage    & 0,998 & 0,992 & \bf{1,000} & \bf{1,000} \\
Shuttle (*) & 0,995 & 0,944 & 0,995 & \bf{0,996} \\
SpamBase    & 0,872 & 0,787 & 0,942 & \bf{0,947} \\
Thyroid     & 0,962 & 0,954 & \bf{0,973} & 0,972 \\
Vertebral   & 0,827 & 0,776 & 0,855 & \bf{0,870} \\
Vowels      & 0,977 & 0,959 & \bf{0,993} & 0,986 \\
WBC         & 0,950 & 0,931 & 0,961 & \bf{0,972} \\
\hline
\end{tabular}

\caption{Results (F1-Score) for ablation study of LEAND, including KDE, AutoEncoder, LEAND with no reconstruction measure and full LEAND.}
\label{ablation}
\end{table}

\section{Conclusions}

This paper presented Quantum Latent Density Estimation for Anomaly Detection (LEAND), a novel method for anomaly detection based on the combination of autoencoders, adaptive Fourier features and density estimation through quantum measurements. This new method was compared against eleven different anomaly detection algorithms, using a framework that included twenty labeled anomaly detection datasets. For each dataset and algorithm, a grid search of the best parameters was performed, and the performance of the winning algorithms was compared using F1-Score (weighted average) as main metric. LEAND showed state-of-the-art performance, being superior than most classic algorithms and comparable to deep learning-based methods. The reliability of the proposed method does not seem to be affected by the features of the dataset, although LEAND highlights in low-dimensional datasets. Also, LEAND performs better than its separate parts (KDE and autoencoder) and there is a noticeable difference when using only reconstruction measures, all of which shows that the combination of density estimation and reconstruction from autoencoders can perform better than these two elements separately.

\bibliography{main_paper}
\bibliographystyle{splncs04}

\end{document}